\documentclass{elsart}
\usepackage{algorithm,algorithmic,amsmath,amssymb,framed,graphicx,multirow}
\usepackage{graphicx,amssymb}
\usepackage{algorithmic}
\usepackage{algorithm}
\usepackage{amsmath}

\usepackage{hyperref}
\usepackage{lineno}

\usepackage{appendix}
\usepackage{longtable}
\usepackage{booktabs}

\usepackage[usenames]{color}
\usepackage{url}
\usepackage{makecell}
\usepackage{enumitem}
\usepackage{comment}

\usepackage{rotating}

\emergencystretch=\maxdimen
\begin{document}
\begin{frontmatter}

\title{A threshold search based memetic algorithm for the disjunctively constrained knapsack problem}
%\author[Angers]{Zequn Wei} \ead{zequn.wei@gmail.com}
\author{Zequn Wei} \ead{zequn.wei@gmail.com}
\and
\author{Jin-Kao Hao\corauthref{cor}}
\corauth[cor]{Corresponding author.} \ead{jin-kao.hao@univ-angers.fr}
\address{LERIA, Universit$\acute{e}$ d'Angers, 2 Boulevard Lavoisier, 49045 Angers, France}
%\address[Angers]{LERIA, Universit$\acute{e}$ d'Angers, 2 Boulevard Lavoisier, 49045 Angers, France}

%\address[IUF]{Institut Universitaire de France, 1 Rue Descartes, 75231  Paris, France}
%\textbf{Submitted x 2020}

\maketitle

%\linenumbers

\begin{abstract}

The disjunctively constrained knapsack problem consists in packing a subset of pairwisely compatible items in a capacity-constrained knapsack such that the total profit of the selected items is maximized while satisfying the knapsack capacity. DCKP has numerous applications and is however computationally challenging (NP-hard). In this work, we present a threshold search based memetic algorithm for solving the DCKP that combines the memetic framework with threshold search to find high quality solutions. Extensive computational assessments on two sets of 6340 benchmark instances in the literature demonstrate that the proposed algorithm is highly competitive compared to the state-of-the-art methods. In particular, we report 24 and 354 improved best-known results (new lower bounds) for Set I (100 instances) and for Set II (6240 instances), respectively. We analyze the key algorithmic components and shed lights on their roles for the performance of the algorithm. The code of our algorithm will be made publicly available.

\end{abstract}

\end{frontmatter}

%\linenumbers

\section{Introduction}
\label{Sec_Intro}

As a generalization of the conventional 0-1 knapsack problem (KP) \cite{kellerer2003knapsack}, the disjunctively constrained knapsack problem (DCKP) is defined as follows. Let $V = \{1,\ldots, n\}$ be a set of $n$ items, where each item $i = \{1,\ldots, n\}$ has a profit $p_i > 0$ and a weight $w_i > 0$. Let $G = (V, E)$ be a conflict graph, where $V$ is the set of $n$ items and an edge $\{i, j\} \in E$ defines the incompatibility of items $i$ and $j$. Let $C > 0$ be the capacity of a given knapsack. Then the DCKP involves finding a subset $S$ of pairwisely compatible items of $V$ to maximize the total profit of $S$ while ensuring that the total weight of $S$ does not surpass the knapsack capacity $C$. Formally, the DCKP can be stated as follows.

\begin{equation}\label{MAX}
(DCKP) \quad \quad \mathrm{Maximize} \quad f(S) = \sum\limits_{i = 1}^n p_ix_i
\end{equation}
\begin{equation}\label{constraint1}
\mathrm{subject \ to} \quad W(S) = \sum\limits_{i = 1}^n w_ix_i \leq C, \ S \subseteq V,
\end{equation}
\begin{equation}\label{constraint2}
x_i + x_j \leq 1, \forall(i, j) \in E,
\end{equation}	
\begin{equation}\label{constraint3}
x_i \in \{0,1\}, i=1,\ldots,n.
\end{equation}

Objective function (\ref{MAX}) commits to maximize the total profit of the selected item set $S$. Constraint (\ref{constraint1}) ensures that the knapsack capacity constraint is satisfied. Constraints (\ref{constraint2}), called disjunctive constraints, guarantee that two incompatible items are never selected simultaneously. Constraints (\ref{constraint3}) force that each item is selected at most once.

It is easy to observe that the DCKP reduces to the NP-hard KP when $G$ is an empty graph. The DCKP is equivalent to the NP-hard maximum weighted independent set problem \cite{johnson1979computers} when the knapsack capacity is unbounded. Moreover, the DCKP is closely related to other combinatorial optimization problems, such as the multiple-choice knapsack problem \cite{kellerer2003knapsack}, and the bin packing problem with conflicts \cite{jansen1999approximation}. In addition to its theoretical significance, the DCKP is a useful model for practical applications where the resources with conflicts cannot be used simultaneously while a given budget envelope cannot be surpassed.

Given the importance of the DCKP, a number of solution methods have been developed including exact, approximation and heuristic algorithms. As the literature review shown in Section \ref{Sec_Rela}, considerable progresses have been continually made since the introduction of the problem. Meanwhile, given the NP-hard nature of the problem, more powerful algorithms are still needed to push the limits of existing methods. 

In this work, we investigate for the first time the population-based memetic framework \cite{moscato1999memetic} for solving the DCKP and design an effective algorithm mixing threshold based local optimization and crossover based solution recombination. The threshold search procedure ensures the main role of search intensification by finding high quality local optimal solutions. The specialized backbone crossover generates promising offspring solutions for search diversification. The algorithm uses also a distance-and-quality strategy for population management. The algorithm has the advantage of avoiding the difficult task of parameter tuning. 

From a perspective of performance assessment, we apply the proposed algorithm to solve the two sets of DCKP benchmark instances in the literature. The results show that for the 100 instances of Set I (optimality still unknown) which were commonly tested by heuristic algorithms, our algorithm discovers 24 new best-known results (new lower bounds) and matches the best-known results for the 76 remaining instances. For the 6240 instances of Set II which were tested by exact algorithms, our algorithm finds 354 improved best lower bounds on the difficult instances whose optimal values are unknown and attains the known optimal results on most of the remaining instances.

The rest of the paper is organized as follows. Section \ref{Sec_Rela} provides a literature review on the DCKP. Section \ref{Sec_TSBMA} presents the proposed algorithm. Section \ref{Sec_Result} shows computational results of our algorithm and provides comparisons with the state-of-the-art algorithms. Section \ref{Sec_AD} analyzes  essential components of the algorithm. Finally, Section \ref{Sec_Conclu} summarizes the work and provides perspectives for future research.

\section{Related work}
\label{Sec_Rela}

The DCKP has attracted considerable attentions in the past two decades. In this section, we review related literature for solving the DCKP. Existing solution methods can be roughly classified into two categories as follows.

\begin{enumerate}
	\item \textit{Exact and approximation algorithms}: These algorithms are able to guarantee the quality of the solutions they find. In 2002, Yamada et al. \cite{yamada2002heuristic} introduced the DCKP and proposed the first implicit enumeration algorithm where the disjunctive constraints are relaxed. In 2007, Hifi and Michrafy \cite{hifi2007reduction} introduced three versions of an exact algorithm based on a local reduction strategy. In 2009, Pferschy and Schauer \cite{pferschy2009knapsack} proposed a pseudo-polynomial time and space algorithm for solving three special cases of the DCKP and proved the DCKP is strongly NP-hard on perfect graphs. In 2016, Salem et al. \cite{salem2018optimization} developed a branch-and-cut algorithm that combines a greedy clique generation procedure with a separation procedure. In 2017, Bettinelli et al. \cite{bettinelli2017branch} presented a branch-and-bound algorithm by combining a upper bounding procedure that considers both the capacity constraint and the disjunctive constraints with a branching procedure that employs a dynamic programming to presolve the 0-1 KP. They generated 4800 DCKP instances with conflict graph densities between 0.1 and 0.9 (see Section \ref{Subsec_Bench}). Also in 2017, Pferschy and Schauer \cite{pferschy2017approximation} applied the approximation methods of modular decompositions and clique separators to the DCKP, and showed complexity results on special graph classes. In 2019, Gurski and Rehs \cite{gurski2019solutions} designed a dynamic programming algorithm and achieved pseudo-polynomial solutions for the DCKP. In 2020, Coniglio et al. \cite{coniglio2020new} presented another branch-and-bound algorithm based on an $n$-ary branching scheme and solved the integer linear programming formulations of the DCKP by the CPLEX solver. They introduced 1440 new and challenging DCKP instances (see Section \ref{Subsec_Bench}).

	\item \textit{Heuristic algorithms}: These algorithms aim to find good near-optimal solutions with a given time. In 2002, Yamada et al. \cite{yamada2002heuristic} proposed a greedy algorithm to generate an initial solution and a 2-opt neighborhood search algorithm to improve the obtained solution. In 2006, Hifi and Michrafy \cite{hifi2006reactive} reported a local search algorithm, which combines a complementary constructive procedure to improve the initial solution and a degrading procedure to diversify the search. They generated a set of 50 DCKP instances with 500 and 1000 items (see Section \ref{Subsec_Bench}), which was widely tested in later studies. In 2012, Hifi and Otmani \cite{hifi2012algorithm} studied two scatter search algorithms. In 2014, Hifi \cite{hifi2014iterative} devised an iterative rounding search-based algorithm that uses a rounding strategy to perform a linear relaxation of the fractional variables. In 2017, Salem et al. \cite{salem2017probabilistic} designed a probabilistic tabu search algorithm (PTS) that operates with multiple neighborhoods. In the same year, Quan and Wu investigated two parallel algorithms: the parallel neighborhood search algorithm (PNS) \cite{quan2017design} and the cooperative parallel adaptive neighborhood search algorithm (CPANS) \cite{quan2017cooperative}. They also designed a new set of 50 DCKP large instances with 1500 and 2000 items (see Section \ref{Subsec_Bench}).
	
\end{enumerate}

Existing studies have significantly contributed to better solving the DCKP. According to the computational results reported in the literature, the parallel neighborhood search algorithm \cite{quan2017design}, the cooperative parallel adaptive neighborhood search algorithm \cite{quan2017cooperative}, and the probabilistic tabu search algorithm \cite{salem2017probabilistic} can be regarded as the state-of-the-art methods for the instances of Set I. For the instances of Set II, the branch-and-bound algorithms presented in \cite{bettinelli2017branch,coniglio2020new} and the integer linear programming formulations solved by the CPLEX solver \cite{coniglio2020new} showed the best performance. 

In this work, we aim to advance the state-of-the-art of solving the problem by proposing the first threshold search based memetic approach, which proves to be effective on the two sets of DCKP instances tested in the literature.

\section{Threshold search based memetic algorithm for the DCKP}
\label{Sec_TSBMA}

Our threshold search based memetic algorithm (TSBMA) for the DCKP is a population-based algorithm combining evolutionary search and local optimization. In this section, we first present the general procedure of the algorithm and then describe its components.

\subsection{General procedure}
\label{Subsec_Intro}

\begin{figure}[h]\centering
\includegraphics[width=3.0in]{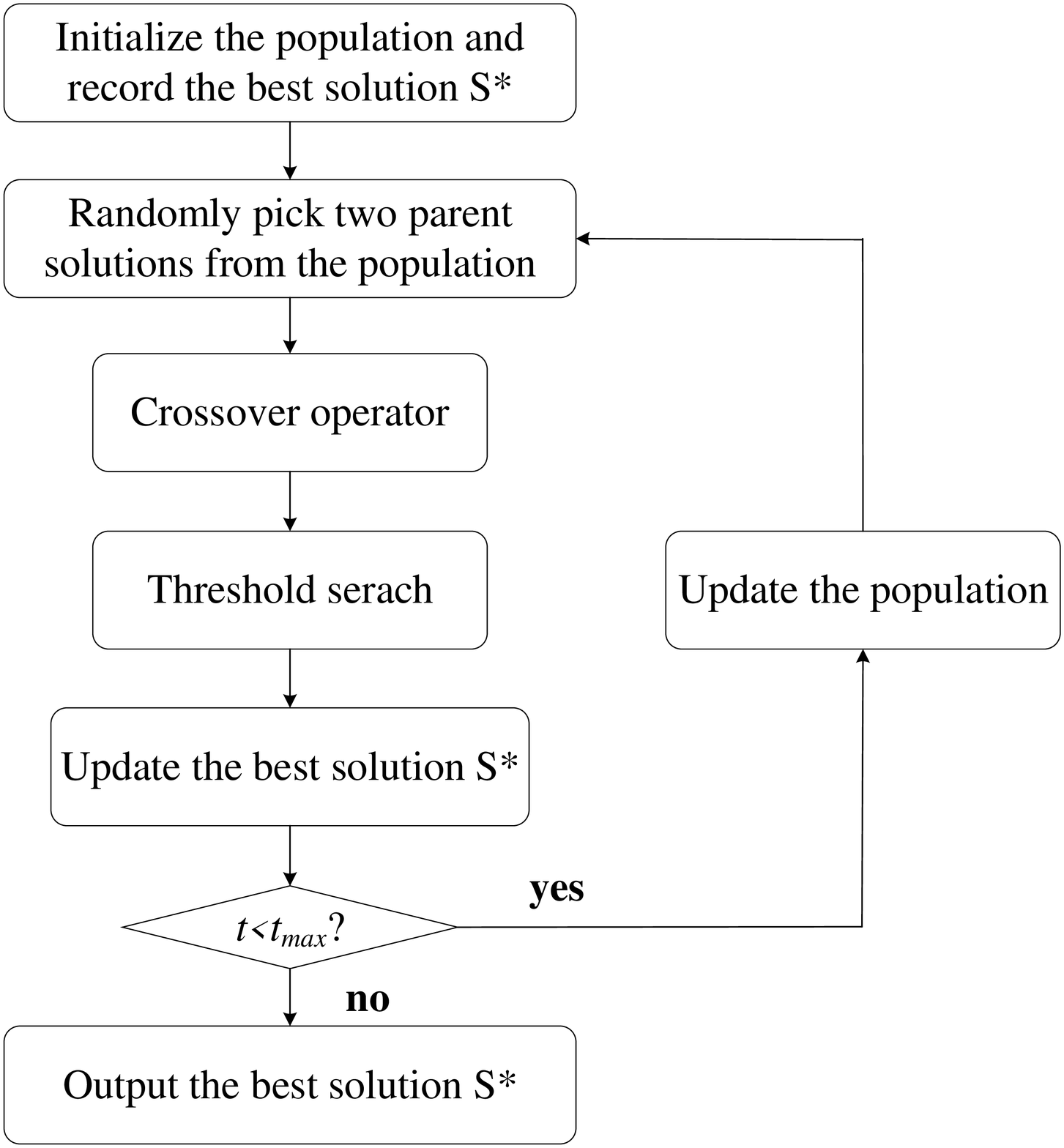}
\caption{Flowchart of the proposed TSBMA algorithm.}
\label{Fig_FC}
\end{figure}

The TSBMA algorithm relies on the general memetic algorithm framework \cite{moscato1999memetic} and follows the design principles recommended in \cite{hao2012memetic}. The flowchart of TSBMA and its pseudo-code are shown in Figure \ref{Fig_FC} and Algorithm \ref{Algo_TSBMA}, respectively. 

The algorithm starts from a set of feasible solutions of good quality that are generated by the population initialization procedure (line 4, Alg. \ref{Algo_TSBMA}, and Section \ref{Subsec_PI}). The best solution is identified and recorded as the overall best solution $S^*$ (line 5, Alg. \ref{Algo_TSBMA}). Then the algorithm enters the main `while' loop (lines 6-15, Alg. \ref{Algo_TSBMA}) to perform a number of generations. At each generation, two solutions are randomly picked and used by the crossover operator to create an offspring solution (line 7-8, Alg. \ref{Algo_TSBMA}, and Section \ref{Subsec_CO}). Afterwards, the threshold search procedure is triggered to perform local optimization with three neighborhoods $N_1$, $N_2$ and $N_3$ (line 9, Alg. \ref{Algo_TSBMA}, and Section \ref{Subsec_TS}). After conditionally updating the overall best solution $S^*$ (line 11-13, Alg. \ref{Algo_TSBMA}), the diversity-based pool updating procedure is applied to decide whether the best solution $S_b$ found during the threshold search should be inserted into the population (line 14, Alg. \ref{Algo_TSBMA}, and Section \ref{Subsec_PU}). Finally, when the given time limit $t_{max}$ is reached, the algorithm returns the overall best solution $S^*$ found during the search and terminates.

\begin{algorithm}
\footnotesize
\caption{Main framework of threshold search based memetic algorithm for the DCKP}\label{Algo_TSBMA}
\begin{algorithmic}[1]
	\STATE \textbf{Input}: Instance $I$, cut-off time $t_{max}$, population $P$, the maximum number of iterations $IterMax$, neighborhoods $N_1$, $N_2$, $N_3$.
	\STATE \textbf{Output}: The overall best solution $S^*$ found.
	\STATE $S^* \gets \emptyset$ \hfill /* Initialize $S^*$ (i.e., $f(S^*) = 0$)*/ 
	\STATE $POP = \{S^1, \ldots, S^{|P|}\} \gets Population\_Initialization(I)$ \hfill /* Section \ref{Subsec_PI} */ 
	\STATE $S^* \gets arg max\{f(S^k)|k = 1, \ldots, p\}$ 
	\WHILE {$Time \leq t_{max}$}
	\STATE Randomly pick two solutions $S^i$ and $S^j$ from the population POP
   	\STATE $S^{o} \gets Crossover\_Operator(S^i, S^j)$ \hfill /* Section \ref{Subsec_CO} */ 
   	\STATE $S_b \gets Threshold\_Search(S^{o}, N_{1-3}, IterMax)$ \hfill /* Section \ref{Subsec_TS} */ 
   	\STATE /* Record the best solution $S_b$ found during threshold search */
   	\IF {$f(S_b) > f(S^*)$}
	\STATE $S^* \gets S_b$ \hfill /* Update the overall best solution $S^*$ found so far */ 
	\ENDIF
	\STATE $POP \gets Pool\_Updating(S_b, POP)$ \hfill /* Section \ref{Subsec_PU} */ 
   	\ENDWHILE   
	\RETURN $S^*$
\end{algorithmic}
\end{algorithm}

\subsection{Solution representation, search space, and evaluation function}
\label{Subsec_SSE}

The DCKP is a subset selection problem. Thus, a candidate solution for a set $V = \{1,\ldots, n\}$ of $n$ items can be conveniently represented by a binary vector $S = (x_1, \ldots, x_n)$, such that $x_i = 1$ if item $i$ is selected, and $x_i = 0$ otherwise. Equivalently, $S$ can also be represented by $S = <A,\bar{A}>$ such that $A = \{q : x_q = 1 \ $in$ \ S\}$ and $\bar{A} = \{p : x_p = 0 \ $in$ \ S\}$.

Let $G = (V, E)$ be the given conflict graph and $C$ be the knapsack capacity. Our TSBMA algorithm explores the following feasible search space $\Omega^F$ satisfying both the disjunctive constraints and the knapsack constraint.

\begin{equation}\label{omega}
\Omega^F = \{x \in \{0,1\}^n : \sum\limits_{i = 1}^n w_ix_i \leq C; x_i + x_j \leq 1, \forall\{i, j\} \in E, 1 \leq i,j \leq n, i \neq j\}
\end{equation}

The quality of a solution $S$ in $\Omega^F$ is determined by the objective value $f(S)$ of the DCKP (Equation \ref{MAX}).

\subsection{Population initialization}
\label{Subsec_PI}

The TSBMA algorithm builds each of the $|P|$ initial solutions of the population $P$ in two steps. First, it randomly adds one by one non-selected items into an individual solution $S^i$ ($i = 1, \ldots, |P|$) until the capacity of the knapsack is reached, while keeping the disjunctive constraints satisfied. Second, to obtain an initial population of reasonable quality, it improves the solution $S^i$ by a short run of the threshold search procedure (Section \ref{Subsec_TS}) by setting $IterMax = 2n$.

It is worth mentioning that the population size $|P|$ is determined according to the number of candidate items $n$ of the given instance, i.e., $|P| = n/100 + 5$. This strategy is based on two considerations. First, since the TSBMA algorithm is powerful enough to solve the instances of small size, a smaller population size can help to reduce the initialization time. Second, the instances of large size are more challenging, a larger population size helps to diversify the search.

\subsection{Local optimization using threshold search}
\label{Subsec_TS}

The local optimization procedure of the TSBMA algorithm relies on the threshold accepting method \cite{DUECK1990161}. To explore a given neighborhood, the method accepts both improving and deteriorating neighbor solutions so long as the solution satisfies a quality threshold. One notices that this method has been successfully applied to solve several knapsack problems (e.g., quadratic multiple knapsack problem \cite{chen2015iterated}, multi-constraint knapsack problem \cite{dueck1991threshold} and multiple-choice knapsack problem \cite{zhou2008algorithm}) and other combinatorial optimization problems (e.g., \cite{castelino1996tabu,tarantilis2004threshold}). In this work, we adopt for the first time this method for solving the DCKP and devise a multiple neighborhood threshold search procedure reinforced by an operation-prohibiting mechanism.

\subsubsection{Main scheme of the threshold search procedure}
\label{Subsec_MTSP}

As shown in Algorithm \ref{Algo_TS}, the threshold search procedure (TSP) starts its process from an input solution and three empty hash vectors (used for the operation-prohibiting mechanism, lines 3-5, Alg. \ref{Algo_TS}). It then performs a number of iterations to explore three neighborhoods (Section \ref{Subsec_MN}) to improve the current solution $S$. Specifically, for each `while' iteration (lines 9-25, Alg. \ref{Algo_TS}), the TSP procedure explores the neighborhoods $N_{1}$, $N_{2}$ and $N_{3}$ in a deterministic way as explained in the next section. Any sampled non-prohibited neighbor solution $S'$ is accepted immediately if the quality threshold $T$ is satisfied (i.e., $f(S') \geq T$). Then the hash vectors are updated for solution prohibition and the best solution found during the TSP procedure is recorded in $S_b$ (line 18-20, Alg. \ref{Algo_TS}). The main search (`while' loop) terminates when 1) no admissible neighbor solution (i.e., non-prohibited and satisfying the quality threshold) exists in the neighborhoods $N_{1}$, $N_{2}$ and $N_{3}$, or 2) the best solution $S_b$ cannot be further improved during $IterMax$ consecutive iterations. Specifically, the quality threshold $T$ is determined adaptively by $f(S_b) - n/10$ ($n$ is the number of items of each instance) while $IterMax$ is set to $(n/500 +5) \times 10000$.

\begin{algorithm}
\footnotesize
\caption{Threshold search procedure}\label{Algo_TS}
\begin{algorithmic}[1]
	\STATE \textbf{Input}: Input solution $S^{o}$, threshold $T$, the maximum number of iterations $IterMax$,  hash vectors $H_1$, $H_2$, $H_3$, hash functions $h_1$, $h_2$, $h_3$, length of hash vectors $L$, neighborhoods $N_1$, $N_2$, $N_3$.
	\STATE \textbf{Output}: The best feasible solution $S_b$ found by threshold search procedure.
	\FOR {$i \gets 0$ \textbf{to} $L-1$}		
	\STATE $H_1[i] \gets 0$; $H_2[i] \gets 0$; $H_3[i] \gets 0$;	\hfill	/* Initialization of hash vectors */
	\ENDFOR
	\STATE $S_b \gets S^{o}$			\hfill	 /* $S_b$ record the best solution found */
	\STATE $S \gets S^{o}$ 	\hfill	 /* $S$ record the current solution */
	\STATE $iter \gets 0$
	\WHILE {$iter \leq IterMax$}
	\STATE Examine the neighborhoods $N_1(S)$, $N_2(S)$, $N_3(S)$ in turn;	/* Section \ref{Subsec_MN} */\\
	 /* Each non-prohibited neighbor solution $S'$ satisfies $H_1[h_1(S{'})] \wedge H_1[h_1(S{'})] = 0 \wedge H_1[h_1(S{'})] = 0$ */	
	\FOR {Each non-prohibited $S'$ of $N_1(S)$ or $N_2(S)$ or $N_3(S)$}
	\IF {$f(S') \geq T$}
	\STATE $S \gets S'$
	\STATE /* Update the hash vectors with $S$, Section \ref{Subsec_OP} */
	\\ $H_1[h_1(S)] \gets 1$; $H_2[h_2(S)] \gets 1$; $H_3[h_3(S)] \gets 1$
	\STATE \textbf{break};
	\ENDIF
    \ENDFOR
    \IF {$f(S) > f(S_b)$}
    \STATE $S_b \gets S$	\hfill	 /* Update the best solution $S_b$ found during threshold search */
    \STATE $iter \gets 0$
    \ELSE
    \STATE $iter \gets iter + 1$
    \ENDIF
	\ENDWHILE
	\RETURN $S_b$
\end{algorithmic}
\end{algorithm}

\subsubsection{Neighborhoods and their exploration}
\label{Subsec_MN}

The TSP procedure examines candidate solutions by exploring three neighborhoods induced by the popular move operators: $add$, $swap$ and $drop$. Let $S$ be the current solution and $mv$ is one of these operators. We use $S' = S \oplus mv$ to denote a feasible neighbor solution obtained by applying $mv$ to $S$ and $N_x$ ($x=1,2,3$) to represent the resulting neighborhoods. To avoid the examination of unpromising neighbor solutions, TSP employs the following dynamic neighborhood filtering strategy inspired by \cite{lai2019two,wei2019iterated}. Let $S'$ be a neighbor solution in the neighborhood currently under examination, and $S_c$ be the best neighbor solution encountered during the current neighborhood examination. Then $S'$ is excluded for consideration if it is no better than $S_c$ (i.e., $f(S') \leq f(S_c)$). By eliminating the unpromising neighbor solutions, TSP increases the efficiency of its neighborhood search. 

Specifically, the associated neighborhoods induced by $add$, $swap$ and $drop$ are defined as follows.

\begin{itemize}[leftmargin = 25pt]
	\item $add(p)$: This move operator expands the selected item set $A$ by one non-selected item $p$ from the set $\bar{A}$ such that the resulting neighbor solution is feasible. This operator induces the neighborhood $N_1$. 
	\begin{equation}\label{N_1}
	N_1(S) = \{S': S'=S \oplus add(p), p \in \bar{A} \}
	\end{equation}
	
	\item $swap(q,p)$: This move operator exchanges a pair of items $(q,p)$ where item $q$ belongs to the selected item set $A$ and $p$ belongs to the non-selected item set $\bar{A}$ such that the resulting neighbor solution is feasible. This operator induces the neighborhood $N_2$.
	\begin{equation}\label{N_2}
	N_2(S) = \{S': S'=S \oplus swap(q,p), q \in A, p \in \bar{A}, f(S') > f(S_c)\}
	\end{equation}
	
	\item $drop(q)$: This operator displaces one selected item $q$ from the set $A$ to the non-selected item set $\bar{A}$ and induces the neighborhood $N_3$.
	\begin{equation}\label{N_3}
	N_3(S) = \{S': S'=S \oplus drop(q), q \in A, f(S') > f(S_c)\}
	\end{equation}
\end{itemize}

One notices that the $add$ operator always leads to a better current solution with an additional eligible item, and thus the neighborhood filtering is not needed for $N_1$. The $drop$ operator always deteriorates the quality of the current solution, and the feasibility of a neighbor solution is always ensured. The $swap$ operator may either increase or decrease the objective value and the feasibility of a neighbor solution needs to be verified. For $N_2$ and $N_3$, neighborhood filtering excludes uninteresting solutions that can in no way be accepted during the TSP process.

The TSP procedure examines the neighborhoods $N_1, N_2,$ and $N_3$ in a token-ring way \cite{di2006neighborhood} to explore different local optimal solutions. For $N_1$, as long as there exists a non-prohibited neighbor solution, TSP selects such a neighbor solution to replace the current solution (ties are broken randomly). Once $N_1$ becomes empty, TSP moves to $N_2$, if there exists a non-prohibited neighbor solution $S'$ satisfying $f(S') \geq T$, TSP selects $S'$ to become the current solution and immediately returns to the neighborhood $N_1$. When $N_2$ becomes empty, TSP continues its search with $N_3$ and explores $N_3$ exactly like with $N_2$. When $N_3$ becomes empty, TSP terminates its search and returns the best solution found $S_b$. TSP may also terminate if its best solution remains unchanged during $IterMax$ consecutive iterations. 

\subsubsection{Operation-prohibiting mechanism}
\label{Subsec_OP}
%, which has shown its effectiveness on other decision-making problems \cite{lai2019two,lai2018solution}

During the TSP procedure, it is important to prevent the search from revisiting a previously encountered solution. For this purpose, TSP utilizes an operation-prohibiting (OP) mechanism that is based on the tabu list strategy \cite{GloverLagunaBook}. To implement the operation-prohibiting (OP) mechanism, we adopt the solution-based tabu search technique \cite{woodruff1993hashing}. Specifically, we employ three hash vectors $H_v$ $(v=1, 2, 3)$ of length $L$ ($|L| = 10^8$) to record previously visited solutions. Given a solution $S =  (x_1, \ldots , x_n)$ ($x_i \in \{0,1\}$), we pre-compute for each item $i$, the weight $\mathcal{W}_i = i^{\gamma_v}$ $(v=1, 2, 3)$, where $\gamma_v$ is equal to $1.2, 1.6, 2.0$, respectively. Then the hash functions $h_v$ {$(v=1, 2, 3)$} are defined as follows.

\begin{equation}\label{Hash_f}
\begin{aligned}
h_v(S) = (\sum_{i=1}^{n} \lfloor \mathcal{W}_i \times x_i \rfloor) \bmod L
\end{aligned}
\end{equation}

The hash value of a neighbor solution $S'$ from the $add$, $swap$ or $drop$ operator can be efficiently computed as follows ($x \in A, y \in \bar{A}$, Section \ref{Subsec_SSE}).

\begin{equation}\label{Hash_f2}
h_v(S') = 
   \begin{cases}
		h(S) + \mathcal{W}_y, & $for\ the$  \ add \ $operator$ \\
		h(S) - \mathcal{W}_x + \mathcal{W}_y, & $for\ the$ \ swap \ $operator$ \\
		h(S) - \mathcal{W}_x, & $for\ the$ \ drop \ $operator$
		\end{cases}
\end{equation}

Starting with the hash vectors set to 0, the corresponding positions in the three hash vectors $H_v$ is updated by 1 whenever a new neighbor solution $S'$ is accepted to replace the current solution $S$ (line 12-16, Alg. \ref{Algo_TS}). For each candidate neighbor solution $S'$, its hashing value $h_v(S')$ is calculated with Equation (\ref{Hash_f2}) in $O(1)$. Then, this neighbor solution $S'$ is previously visited if $H_1[h_1(S')] \wedge H_2[h_2(S')] \wedge H_3[h_3(S')] = 1$ and is prohibited from consideration by the TSP procedure.

\subsection{Crossover operator}
\label{Subsec_CO}

The crossover operator generally creates new solutions by recombining two existing solutions. For the DCKP, we adopt the idea of the double backbone-based crossover (DBC) operator \cite{zhou2018memetic} and adapt it to the problem.

Given two solutions $S^i$ and $S^j$, we use them to divide the set of $n$ items into three subsets: the common items set $X_1 = S^i \cap S^j$, the unique items set $X_2 = (S^i \cup S^j) \setminus (S^i \cap S^j)$ and the unrelated set $X_3 = V \setminus (S^i \cup S^j)$. The basic idea of the DBC operator is to generate an offspring solution $S^{o}$ by selecting all items in set $X_1$ (the first backbone) and some items in set $X_2$ (the second backbone), while excluding items in set $X_3$.

As shown in Algorithm \ref{Algo_CO}, from two randomly selected parent solutions $S^i$ and $S^j$, the DBC operator generates $S^{o}$ in three steps. First, we initialize $S^{o}$ by setting all the variables $x_a^{o}$ ($a = 1, \ldots, n$) to 0 (line 3, Alg. \ref{Algo_CO}). Second, we identify the common items set $X_1$ and the unique items set $X_2$ (line 4-10, Alg. \ref{Algo_CO}). Third, we add all items belonging to $X_1$ into $S^{o}$ and randomly add items from $X_2$ into $S^{o}$ until the knapsack constraint is reached (line 11-17, Alg. \ref{Algo_CO}). Note that the knapsack and disjunctive constraints are always satisfied during the crossover process.

Since the DCKP is a constrained problem, the DBC operator adopted for TSBMA has several special features to handle the constraints, which is different from the DBC operator introduced in \cite{zhou2018memetic}. First, we iteratively add an item into $S^{o}$ by selecting one item from the unique items set $X_2$ randomly until the knapsack constraint is reached, while each item in $X_2$ is considered with a probability $p_0$ ($0 < p_0 < 1$) in \cite{zhou2018memetic}. Second, unlike \cite{zhou2018memetic} where a repair operation is used to achieve a feasible offspring solution, our DBC operator ensures the satisfaction of the problem constraints during the offspring generation process.

\begin{algorithm}
\footnotesize
\caption{The double backbone-based crossover operator}\label{Algo_CO}
\begin{algorithmic}[1]
	\STATE \textbf{Input}: Two parent solutions $S^i = (x_1^i, x_2^i, \ldots, x_n^i)$ and $S^j = (x_1^j, x_2^j, \ldots, x_n^j)$.
	\STATE \textbf{Output}: An offspring solution $S^{o} = (x_1^{o}, x_2^{o}, \ldots, x_n^{o})$.
	\STATE $S^{o} \gets \emptyset$ \hfill /* Initialize $S^{o}$ (i.e., $f(S^{o}) = 0$)*/
	\FOR {$a \gets 1$ \textbf{to} $n$}
	\IF {$x_a^i = 1$ and $x_a^j = 1$}
	\STATE $X_1 \gets a$		 \hfill /* $X_1$ is the common items set */%/* The first backbone */
	\ELSIF {$x_a^i = 1$ or $x_a^j = 1$} 
	\STATE	$X_2 \gets a$	 \hfill /* $X_2$ is the unique items set */
	\ENDIF
    \ENDFOR
    \STATE $S^{o} \gets X_1$  \hfill /* Add all items belonging to $X_1$ into $S^{o}$ */
    \STATE Randomly shuffle all items in $X_2$;
	\FOR {each $a \in$ $X_2$}
	\IF {$S^{o} \cup (x_a^{o} = 1)$ is a feasible solution}
	\STATE $x_a^{o} \gets 1$	 \hfill /* The second backbone */
	\ENDIF
    \ENDFOR
	\RETURN $S^{o}$
\end{algorithmic}
\end{algorithm}

\subsection{Population updating}
\label{Subsec_PU}

Once a new offspring solution is obtained by the DBC operator in the last section, it is further improved by the threshold search procedure presented in Section \ref{Subsec_TS}. Then we adopt a diversity-based population updating strategy \cite{lai2018two} to decide whether the improved offspring solution should replace an existing solution in the population. This strategy is beneficial to balance the quality of the offspring solution and its distance from the population.

To accomplish this task, we temporarily insert the improved offspring solution into the population and compute the distance (Hamming distance) between any two solutions in the population. Then we obtain the goodness score of each solution in the same way as proposed in \cite{lai2018two}. Finally, the worst solution in the population is identified according to the goodness score and deleted from the population.

\subsection{Time complexity of TSBMA}
\label{Subsec_TC}

As shown in Section \ref{Subsec_PI}, the population initialization procedure includes two steps. Given a DCKP instance with $n$ items, the first step of random selection takes time $O(n)$. Given an input solution $S=<A,\bar{A}>$ (see Section \ref{Subsec_SSE}), the complexity of one iteration of the TSP procedure is $O((n + |A| \times |\bar{A}|))$. Then the second step of the initialization procedure can be realized in $O([(n + |A| \times |\bar{A}|)] \times IterMax)$, where $IterMax$ is set to $2n$ in the initialization procedure. The complexity of the population initialization procedure is $O(n^3)$.

Now we consider the four procedures in the main loop of the TSBMA algorithm: parent selection, crossover operator, the TSP procedure and population updating. The parent selection procedure is realized in $O(1)$. The crossover operator takes time $O(n)$. The complexity of the TSP procedure is $O([(n + |A| \times |\bar{A}|)] \times IterMax)$, where $IterMax$ is determined in Section \ref{Subsec_MTSP}. The population updating procedure can be achieved in $O(n|P|)$, where $|P|$ is the population size. Then, the complexity of one iteration of the main loop of the TSBMA algorithm is $O(n^2 \times IterMax)$.

\section{Computational results and comparisons}
\label{Sec_Result}

In this section, we assess the proposed TSBMA algorithm by performing extensive experiments and making comparisons with state-of-the-art DCKP algorithms. We report computational results on two sets of 6340 benchmark instances.

\subsection{Benchmark instances}
\label{Subsec_Bench}

The benchmark instances of the DCKP tested in our experiments were widely used in the literature, which can be divided into two sets (see Tables \ref{Intro_insI} and \ref{Intro_insII} for the main characteristics of these instances).

\textbf{Set I (100 instances)}: These instances are grouped into 20 classes (each with 5 instances) and named by $xIy$ ($x = \{1,\ldots, 20\}$ and $y = \{1,\ldots, 5\}$). The first 50 instances ($1Iy$ to $10Iy$) were introduced in 2006 \cite{hifi2006reactive} and have the following features: number of items $n=500$ or $1000$, capacity $C=1800$ or $2000$, and density $\eta$ going from 0.05 to 0.40. Note that the density is given by $2m / n(n-1)$, where $m$ is the number of disjunctive constraints (i.e., the number of edges of the conflict graph). These instances have an item weight $w_i$ uniformly distributed in $[1,100]$ and a profit $p_i = w_i + 10$. For the instance classes $11Iy$ to $20Iy$ introduced in 2017 \cite{quan2017cooperative}, the number of items $n$ is set to 1500 or 2000, the capacity $C$ is set to 4000, and the density $\eta$ ranges from 0.04 to 0.20. These instances have an item weight $w_i$ uniformly distributed in $[1,400]$ and a profit $p_i$ equaling $w_i + 10$.

\textbf{Set II (6240 instances)}: This set of instances was introduced in 2017 \cite{bettinelli2017branch} and expanded in 2020 \cite{coniglio2020new}. For the four correlated instance classes $C1$ to $C15$ (denoted by $CC$) and four random classes $R1$ to $R15$ (denoted by $CR$), the number of items $n$ is from 60 to 1000, the capacity $C$ is from 150 to 15000, and the density $\eta$ is from 0.10 to 0.90. Each of these eight classes contains 720 instances. For the correlated instance class $SC$ and the random instance class $SR$ of the sparse graphs, the number of items $n$ is from 500 to 1000, the capacity $C$ is from 1000 to 2000, and the density $\eta$ is from 0.001 to 0.05. Each of these two classes contains 240 DCKP instances. More details about this set of instances can be found in \cite{coniglio2020new}.

\renewcommand{\baselinestretch}{1}\large
\begin{table}[!htbp]\centering
	\caption{Summary of main characteristics of the 100 DCKP instances of Set I.} 
	\begin{scriptsize}
	\setlength{\tabcolsep}{4mm}{
	\begin{tabular}{llllllllll}
	\toprule[0.75pt]

Class & Total & $n$ & $C$ & $\eta$ & Class & Total & $n$ & $C$ & $\eta$  \\
\hline

$1Iy$ & 5 & 500 & 1800 & 0.10 & $11Iy$ & 5 & 1500 & 4000 & 0.04 \\
$2Iy$ & 5 & 500 & 1800 & 0.20 & $12Iy$ & 5 & 1500 & 4000 & 0.08 \\      
$3Iy$ & 5 & 500 & 1800 & 0.30 & $13Iy$ & 5 & 1500 & 4000 & 0.12 \\
$4Iy$ & 5 & 500 & 1800 & 0.40 & $14Iy$ & 5 & 1500 & 4000 & 0.16 \\
$5Iy$ & 5 & 1000 & 1800 & 0.05 & $15Iy$ & 5 & 1500 & 4000 & 0.20 \\
$6Iy$ & 5 & 1000 & 2000 & 0.06 & $16Iy$ & 5 & 2000 & 4000 & 0.04 \\      
$7Iy$ & 5 & 1000 & 2000 & 0.07 & $17Iy$ & 5 & 2000 & 4000 & 0.08 \\
$8Iy$ & 5 & 1000 & 2000 & 0.08 & $18Iy$ & 5 & 2000 & 4000 & 0.12 \\    
$9Iy$ & 5 & 1000 & 2000 & 0.09 & $19Iy$ & 5 & 2000 & 4000 & 0.16 \\
$10Iy$ & 5 & 1000 & 2000 & 0.10 & $20Iy$ & 5 & 2000 & 4000 & 0.20 \\     

\bottomrule[0.75pt]
\end{tabular}}
\label{Intro_insI}
\end{scriptsize}
\end{table} 
\renewcommand{\baselinestretch}{1}\large\normalsize

\renewcommand{\baselinestretch}{1}\large
\begin{table}[!htbp]\centering
	\caption{Summary of main characteristics of the 6240 DCKP instances of Set II.} 
	\begin{scriptsize}
	\setlength{\tabcolsep}{4mm}{
	\begin{tabular}{llllllll}
	\toprule[0.75pt]

\multirow{2}{*}{Class} &\multirow{2}{*}{Total} & \multicolumn{2}{c}{$n$} & \multicolumn{2}{c}{$C$} & \multicolumn{2}{c}{$\eta$} \\
\cmidrule(lr){3-4} \cmidrule(lr){5-6} \cmidrule(lr){7-8}
  & & Min & Max & Min & Max & Min & Max \\  
\hline

$C1$ & 720 & 60 & 1000 & 150 & 1000 & 0.10 & 0.90 \\
$C3$ & 720 & 60 & 1000 & 450 & 3000 & 0.10 & 0.90 \\
$C10$ & 720 & 60 & 1000 & 1500 & 10000 & 0.10 & 0.90 \\
$C15$ & 720 & 60 & 1000 & 15000 & 15000 & 0.10 & 0.90 \\
$R1$ & 720 & 60 & 1000 & 150 & 1000 & 0.10 & 0.90 \\
$R3$ & 720 & 60 & 1000 & 450 & 3000 & 0.10 & 0.90 \\
$R10$ & 720 & 60 & 1000 & 1500 & 10000 & 0.10 & 0.90 \\
$R15$ & 720 & 60 & 1000 & 15000 & 15000 & 0.10 & 0.90 \\
$SC$ & 240 & 500 & 1000 & 1000 & 2000 & 0.001 & 0.05 \\
$SR$ & 240 & 500 & 1000 & 1000 & 2000 & 0.001 & 0.05 \\

\bottomrule[0.75pt]
\end{tabular}}
\label{Intro_insII}
\end{scriptsize}
\end{table} 
\renewcommand{\baselinestretch}{1}\large\normalsize

\subsection{Experimental settings}
\label{Subsec_Setting}

\textbf{Reference algorithms.} For the 100 DCKP instances of Set I that were widely tested by heuristic algorithms, we adopt as our reference methods three state-of-the-art heuristic algorithms: parallel neighborhood search algorithm (PNS) \cite{quan2017design}, cooperative parallel adaptive neighborhood search algorithm (CPANS) \cite{quan2017cooperative}, and probabilistic tabu search algorithm (PTS) \cite{salem2017probabilistic}. Note that PTS only reported results of the 50 instances $1Iy$ to $10Iy$, since the other 50 instances of $11Iy$ to $20Iy$ were designed later. For the 6240 DCKP instances of Set II that were only tested by exact algorithms until now, we cite the results of three best performing methods: branch-and-bound algorithms BCM \cite{bettinelli2017branch} and CFS \cite{coniglio2020new}) as well as the integer linear programming formulations solved by the CPLEX solver (ILP) \cite{coniglio2020new}. 

\textbf{Computing platform.} The proposed TSBMA algorithm was written in C++\footnote{The code of our TSBMA algorithm will be available at: \url{http://www.info.univ-angers.fr/pub/hao/DCKP_TSBMA.html.}\label{foot}} and compiled using the g++ compiler with the -O3 option. All experiments were carried out on an Intel Xeon E5-2670 processor (2.5 GHz CPU and 2 GB RAM) under the Linux operating system. The results of the main reference algorithms have been obtained on computing platforms with the following features: an Intel Xeon processor with 2$\times$3.06 GHz for CPANS and PNS, an Intel Pentium i5-6500 processor with 3.2 GHz and 4 GB RAM for PTS, and an Intel Xeon E5-2695 processor with 3.00GHz for CFS. Note that the parallel algorithms PNS and CPANS used 10 to 400 processors to obtain the results.

\textbf{Parameter settings.}  The TSBMA algorithm does not require parameter tuning (it is parameter-free). However, for the 6240 instances of Set II (with a wide range of densities and number of items), we adjusted the threshold $T$ (see Section \ref{Subsec_MTSP}) to $T = MinP + rand(20)$, where $MinP$ is the minimum profit value for each instance tested.

\textbf{Stopping condition.} For the 100 DCKP instances of Set I, the TSBMA algorithm adopted the same cut-off time as the reference algorithms (PNS, CPANS and PTS), i.e., 1000 seconds. Note that for the instances $11Iy$ to $20Iy$, PNS used a much longer limit of 2000 seconds. Given its stochastic nature, TSBMA was performed 20 times independently with different random seeds to solve each instance. For the 6240 instances of Set II, the cut-off time was set to 600 seconds as in the CFS algorithm and the number of repeated runs was set to 10.

\subsection{Computational results and comparisons}
\label{Subsec_Result}

In this section, we first present summarized comparisons of the proposed TSBMA algorithm against each reference algorithm on the 100 instances of Set I, and then show the comparative results on the 6240 DCKP instances of Set II. The detailed computational results of our algorithm and the reference algorithms on the instances of Set I are shown in the Appendix, while our solution certificates for these 100 instances are available at the webpage indicated in footnote \ref{foot}. For the 6240 instances of Set II, we report their objective values at the same website.

\subsubsection{Comparative results on the 100 benchmark instances of Set I}
\label{Compara_I}

The comparative results of the TSBMA algorithm and each reference algorithm are summarized in Table \ref{Summary_insI}. Column 1 indicates the pairs of compared algorithms and column 2 gives the names of instance class. Column 3 shows the quality indicators: the best objective value ($f_{best}$) and the average objective value ($f_{avg}$) (when the average results are available in the literature). The following columns \#Wins, \#Ties and \#Losses present the number of instances for which TSBMA achieves a better, equal and worse result according to the indicators. To further analyze the performance of our algorithm, we carried out the Wilcoxon signed-rank test to verify the statistical significance of the compared results between TSBMA and each compared algorithm in terms of the $f_{best}$ and $f_{avg}$ values (when the average results are available in the literature). The outcomes of the Wilcoxon tests are shown in the last column where `NA' means that the two sets of compared results are exactly the same.

From Table \ref{Summary_insI}, one observes that the TSBMA algorithm competes very favorably with all the reference algorithms by reporting improved or equal results on all the instances. Compared to the probabilistic tabu search algorithm (PTS) \cite{salem2017probabilistic} which reported results only on the first 50 instances of classes $1Iy$ to $10Iy$, TSBMA finds 8 (45) better $f_{best}$ ($f_{avg}$) values, while matching the remaining results. Compared to the two parallel algorithms (PNS) \cite{quan2017design} and (CPANS) \cite{quan2017cooperative} that reported only the $f_{best}$ values, TSBMA obtained 35 and 29 better $f_{best}$ results, respectively. The small $p$-$values$ ($< 0.05$) from the Wilcoxon tests between TSBMA and its competitors indicate that the performance differences are statistically significant. Finally, it is remarkable that our TSBMA algorithm discovered 24 new lower bounds on the instances $11Iy$ to $20Iy$ (see the detailed results shown in the Appendix). 

\begin{table}[!htp]\centering
	\caption{Summarized comparisons of the TSBMA algorithm against each reference algorithm with the $p$-$values$ of the Wilcoxon signed-rank test on the 100 DCKP instances of Set I.}
	\renewcommand{\baselinestretch}{1.0}\large\normalsize
	\begin{scriptsize}
	\setlength{\tabcolsep}{1mm}{
	\begin{tabular}{l|cccccc}
	\toprule[0.75pt]
  Algorithm pair & Instance & Indicator & \#Wins & \#Ties & \#Losses & $p$-$value$ \\
\hline

TSBMA vs. PTS \cite{salem2017probabilistic}  & $1Iy - 10Iy$ (50) & $f_{best}$ & 8 & 42 & 0 & 1.40e-2 \\
	&		& $f_{avg}$ & 45 & 5 & 0 & 5.34e-9 \\
\hline
TSBMA vs. PNS \cite{quan2017design} & $1Iy - 10Iy$ (50) & $f_{best}$ & 9 & 41 & 0 & 8.91e-3 \\
	& $11Iy - 20Iy$ (50) & $f_{best}$ & 26 & 24 & 0 & 8.25e-6 \\
\hline
TSBMA vs. CPANS \cite{quan2017cooperative} & $1Iy - 10Iy$ (50) & $f_{best}$ & 0 & 50 & 0 & NA \\
	& $11Iy - 20Iy$ (50) & $f_{best}$ & 29 & 21 & 0 & 2.59e-6 \\

	\bottomrule[0.75pt]
	\end{tabular}}
\label{Summary_insI}
\end{scriptsize}
\end{table} 
\renewcommand{\baselinestretch}{1.0}\large\normalsize

\begin{figure}[H]
\begin{minipage}{0.44\textwidth} 
\includegraphics[width=3.1in]{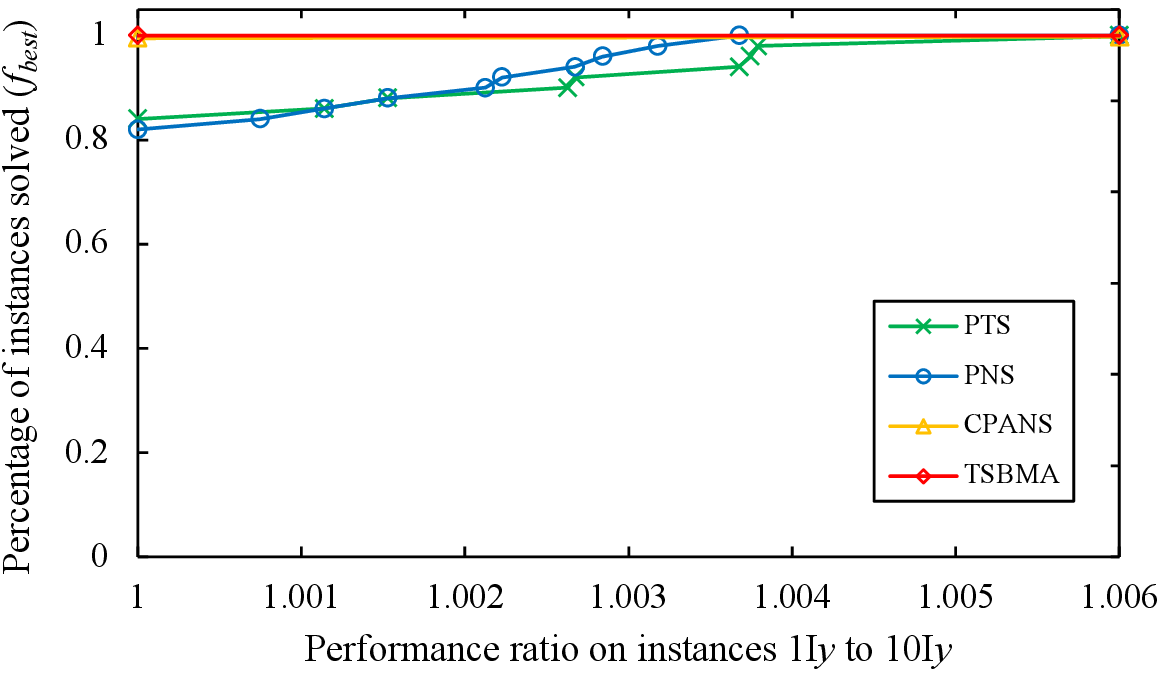}
\end{minipage}
\hfill 
\begin{minipage}{0.44\textwidth} 
\includegraphics[width=3.1in]{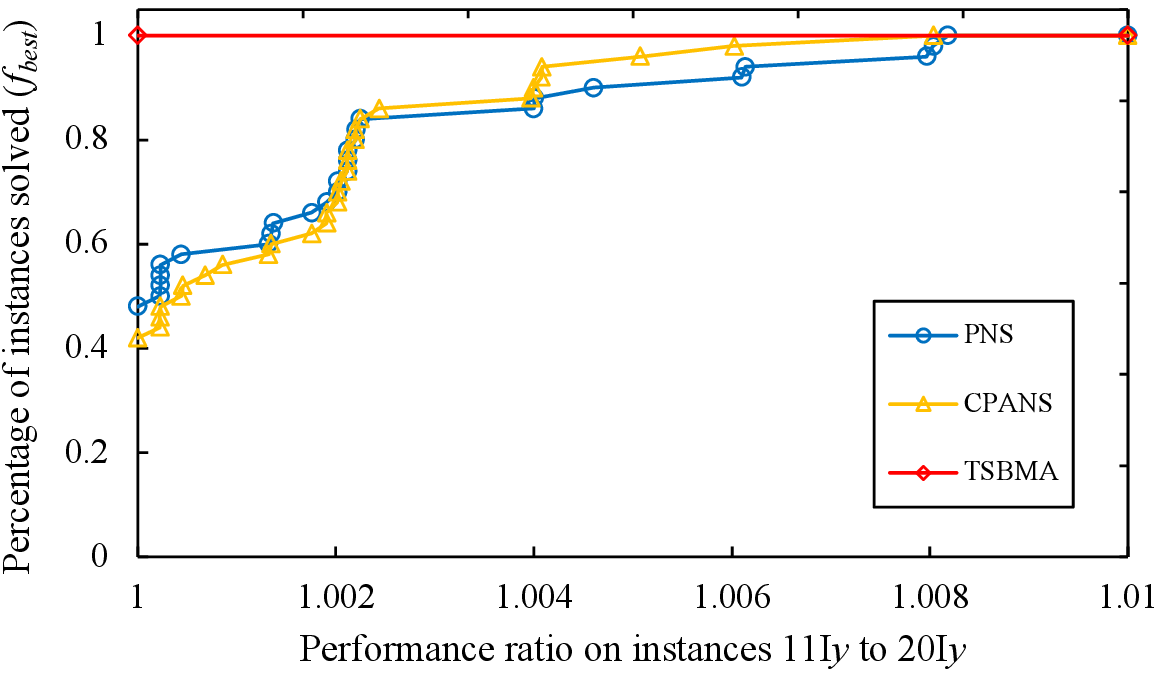}
\end{minipage}

\caption{Performance profiles of the compared algorithms on the 100 DCKP instances of Set I.}
\label{Fig_PP}
\end{figure}

To complete the assessment, we provide the performance profiles \cite{dolan2002benchmarking} of the four compared algorithms on the 100 instances of Set I. Basically, the performance profile of an algorithm shows the cumulative distribution for a given performance metric, which reveals the overall performance of the algorithm on a set of instances. In our case, the plots concern the best objective values ($f_{best}$) of the compared algorithms since the average results of some reference algorithms are not available in the literature. Given a set of algorithms (solvers) $\mathcal{S}$ and an instance set $\mathcal{P}$, the performance ratio is given by $r_{p,s} = \frac{f_{p,s}}{min\{f_{p,s} : s \in \mathcal{S} \}}$, where $f_{p,s}$ is the $f_{best}$ value of instance $p$ of $\mathcal{P}$ obtained by algorithm $s$ of $\mathcal{S}$. The performance profiles are shown in Figure \ref{Fig_PP}, where the performance ratio and the percentage of instances solved by each compared algorithm are displayed on the $X$-$axis$ and $Y$-$axis$, respectively. When the value of $X$-$axis$ is 1, the corresponding value of $Y$-$axis$ indicates the fraction of instances for which algorithm $s$ can reach the best $f_{best}$ value of the set $\mathcal{S}$ of the compared algorithms.

From Figure \ref{Fig_PP}, we observe that our TSBMA algorithm has a very good performance on the 100 benchmark instances of Set I compared to the reference algorithms. For the 50 instances $1Iy$ to $10Iy$, TSBMA and CPANS are able to reach 100\% best $f_{best}$ values on these 50 instances, while PTS and PNS fail on around 15\% of the instances. When considering the 50 instances $11Iy$ to $20Iy$, the plot of TSBMA strictly runs above the plots of PNS and CPANS, revealing that our algorithm dominates the reference algorithms on these 50 instances. These outcomes again confirm the high performance of our TSBMA algorithm.

\subsubsection{Comparative results on the 6240 benchmark instances of Set II}
\label{Compara_II}

Table \ref{Summary_insII} summarizes the comparative results of our TSBMA algorithm on the 6240 instances of Set II, together with the three reference algorithms mentioned in \ref{Subsec_Setting}. Note that three ILP formulations were studied in \cite{coniglio2020new}, we extracted the best results of these formulations in Table \ref{Summary_insII}, i.e., the results on instances $CC$ and $CR$ (conflict graph density from 0.10 to 0.90) with ILP$_2$ and the results on very sparse instances $SC$ and $SR$ (conflict graph density from 0.0001 to 0.005) with ILP$_1$. Columns 1 and 2 of Table \ref{Summary_insII} identify each instance class and the total number of instances of the class. Columns 3 to 5 indicate the number of instances solved to optimality by the three reference algorithms. Column 6 shows the number of instances for which our TSBMA algorithm reaches the optimal solution proved by exact algorithms. The number of new lower bounds (denoted by NEW LB in Table \ref{Summary_insII}) found by TSBMA is provided in column 7. The best results of the compared algorithms are highlighted in bold. In order to further evaluate the performance of our algorithm, we summarize the available comparative results between MSBTS and the main reference algorithm CFS in columns 8 to 10. The last three rows provide an additional summary of the results for each column.

From Table \ref{Summary_insII}, we observe that TSBMA performs globally very well on the instances of Set II. For the 5760 $CC$ and $CR$ instances, TSBMA reaches most of the proved optimal solutions (5381 out of 5389) and discovers new lower bounds for 323 difficult instances whose optima are still unknown. For the 240 very sparse $SC$ instances, TSBMA matches 195 out of 200 proved optimal solutions and finds 24 new lower bounds for the remaining instances. Although TSBMA successfully solves only 9 out of the 229 solved very sparse $SR$ instances, it discovers 7 new lower bounds. The high performance of TSBMA is further evidenced with the comparison with the best exact algorithm CFS (last three columns).

Notice that the performance of CPLEX with ILP$_1$ is better than TSBMA as well as the two reference algorithms BCM and CFS on the two classes of very sparse instances ($SC$ and $SR$). As analyzed in \cite{coniglio2020new}, one of the main reasons is that the LP relaxation of ILP$_1$ provides a very strong upper bound, which makes the ILP$_1$ formulation very suitable for solving very sparse instances. The disjunctive constraints become very weak when the conflict graph is very sparse. For these two classes of instances, the pure branch-and-bound CFS algorithm is more effective on extremely sparse instances with densities up to 0.005. On the contrary, our TSBMA algorithm is more suitable for solving sparse instances with densities between 0.01 and 0.05. In fact, the new lower bounds found by TSBMA all concern instances with a density of 0.05. Finally, the TSBMA algorithm remains competitive on the 240 correlated sparse instances $SC$, even if the density is the smallest (0.001), which means that only the random sparse instance class $SR$ is challenging for TSBMA.

In summary, our TSBMA algorithm is computational efficient on a majority of the 6240 benchmark instances of Set II and is able to discover new lower bounds on 354 difficult DCKP instances, whose optimal solutions are still unknown.

\renewcommand{\baselinestretch}{1}\large
\begin{table}[!htbp]\centering
	\caption{Summarized comparisons of the TSBMA algorithm against each reference algorithm on the 6240 DCKP instances of Set II.} 
	\begin{scriptsize}
	\setlength{\tabcolsep}{1.2mm}{
	\begin{tabular}{llllllllll}
	\toprule[0.75pt]

\multirow{2}{*}{Class} & \multirow{2}{*}{Total} & \multicolumn{1}{c}{ILP$_{1,2}$ \cite{coniglio2020new}} & \multicolumn{1}{c}{BCM \cite{bettinelli2017branch}} & \multicolumn{1}{c}{CFS \cite{coniglio2020new}} & \multicolumn{2}{c}{TSBMA (this work)} & \multicolumn{3}{c}{TSBMA vs. CFS} \\
\cmidrule(lr){3-3} \cmidrule(lr){4-4} \cmidrule(lr){5-5} \cmidrule(lr){6-7} \cmidrule(lr){8-10} 
  & & Solved & Solved & Solved & Solved & New LB & \#Wins & \#Ties & \#Losses  \\  
\hline

$C1$ & 720 & \textbf{720} & \textbf{720} & \textbf{720} & \textbf{720} & 0 & 0 & 720 & 0 \\
$C3$ & 720 & 584 & \textbf{720} & \textbf{720} & 716 & 0 & 0 & 716 & 4 \\
$C10$ & 720 & 446 & 552 & \textbf{617} & \textbf{617} & \textbf{91} & 91 & 629 & 0 \\
$C15$ & 720 & 428 & 550 & \textbf{600} & \textbf{600} & \textbf{117} & 117 & 603 & 0 \\
$R1$ & 720 & \textbf{720} & \textbf{720} & \textbf{720} & 717 & 0 & 0 & 717 & 3 \\
$R3$ & 720 & 680 & \textbf{720} & \textbf{720} & \textbf{720} & 0 & 0 & 720 & 0 \\
$R10$ & 720 & 508 & 630 & \textbf{670} & 669 & \textbf{37} & 37 & 681 & 2 \\
$R15$ & 720 & 483 & 590 & \textbf{622} & \textbf{622} & \textbf{78} & 78 & 641 & 1 \\
$SC$ & 240 & \textbf{200} & 109 & 156 & 195 & \textbf{24} & 70 & 165 & 5 \\
$SR$ & 240 & 229 & 154 & 176 & 9 & \textbf{7} & 43 & 8 & 189 \\
\hline
Total on $CC$ and $CR$  & 5760 & 4569 & 5201 & 5389 & 5381 & 323 & 323 & 5427 & 10 \\
Total on $SC$ and $SR$  & 480 & 429 & 263 & 332 & 204 & 31 & 113 & 173 & 194 \\
\hline
Grand total & 6240 & 4998 & 5424 & 5721 & 5585 & 354 & 436 & 5600 & 204 \\

\bottomrule[0.75pt]
\end{tabular}}
\label{Summary_insII}
\end{scriptsize}
\end{table} 
\renewcommand{\baselinestretch}{1}\large\normalsize

\section{Analysis and discussions}
\label{Sec_AD}

In this section, we analyze two essential components of the TSBMA algorithm: the importance of the threshold search and the contribution of the operation-prohibiting mechanism. The studies in this section are based on the 50 benchmark instances $11Iy$ to $20Iy$ of Set I.

\subsection{Importance of the threshold search}
\label{Analy_TS}

\begin{figure}[H]
\begin{minipage}{0.44\textwidth} 
\includegraphics[width=3.1in]{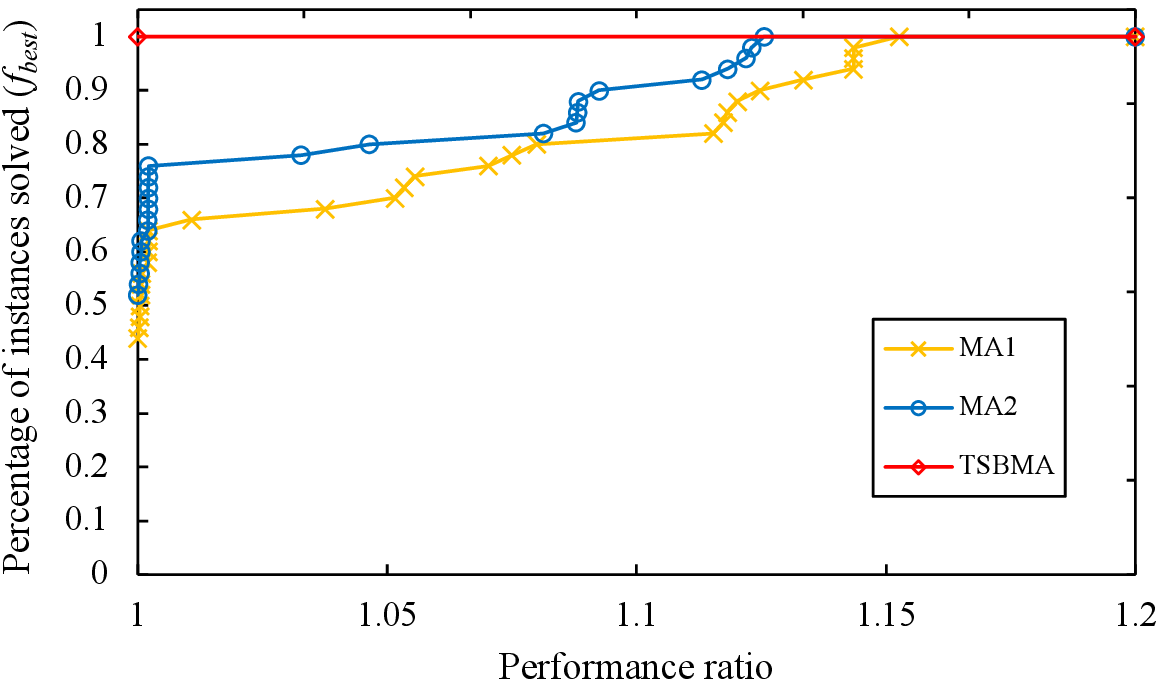}
\end{minipage}
\hfill 
\begin{minipage}{0.44\textwidth} 
\includegraphics[width=3.1in]{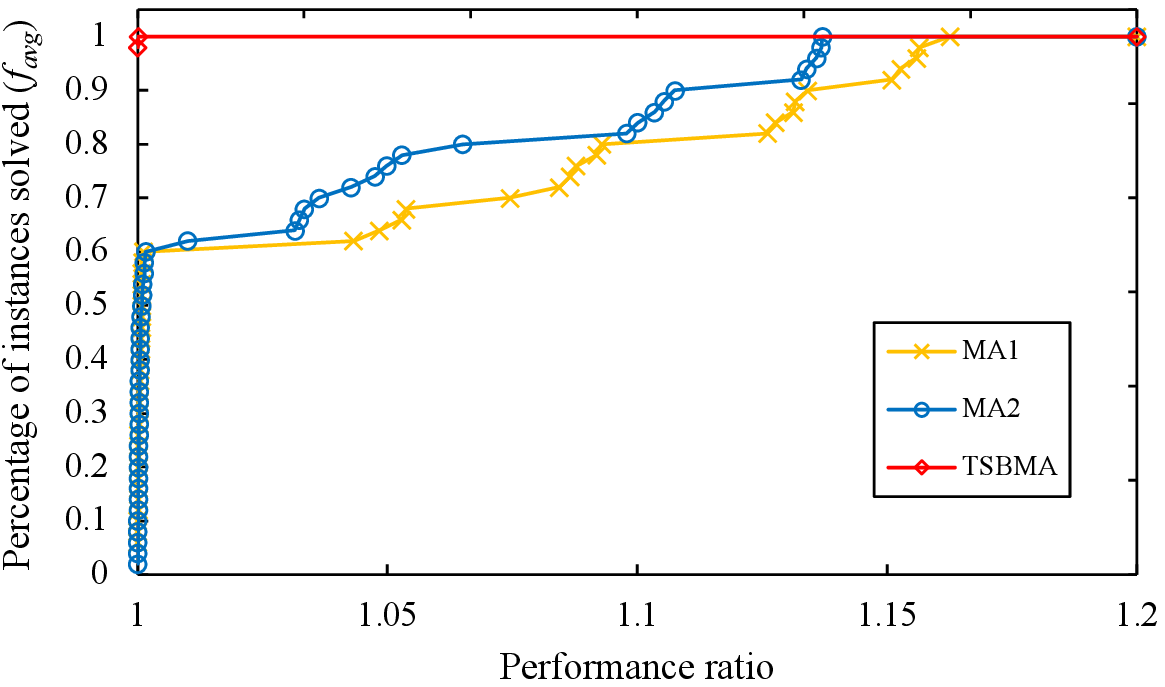}
\end{minipage}

\caption{Performance profiles of the compared algorithms on the 50 DCKP instances $11Iy$ to $20Iy$.}
\label{Fig_PP2}
\end{figure}

The threshold search procedure of the TSBMA algorithm is the first adaptation of the threshold accepting method to the DCKP. To assess the importance of this component, we compare TSBMA with two TSBMA variants by replacing the TSP procedure with the $first$-$improvement$ descent procedure and $best$-$improvement$ descent procedure. In other words, these variants (named as MA1 and MA2) use, in each iteration, the first and the best improving solution $S'$ in the neighborhood to replace the current solution, respectively. We carried out an experiment by running the two variants to solve the 50 instances $11Iy$ to $20Iy$ with the same experimental settings of Section \ref{Subsec_Setting}. The performance profiles of TSBMA and these TSBMA variants are shown in Figure \ref{Fig_PP2} based on the best objective values (left sub-figure) and  the average objective values (right sub-figure).

From Figure \ref{Fig_PP2}, we can clearly observe that TSBMA dominates MA1 and MA2 according to the cumulative probability obtained by the $f_{best}$ and $f_{avg}$ values. The plots of TSBMA strictly run above the plots of MA1 and MA2, indicating TSBMA performs always better than the two variants. This experiment implies that the adopted threshold search procedure of TSBMA is relevant for its performance.

\subsection{Contribution of the operation-prohibiting mechanism}
\label{Analy_OP}

\renewcommand{\baselinestretch}{0.8}\large\normalsize
\begin{table}[!htbp]
\centering
	\caption{Comparison between TSBMA$^-$ (without the OP mechanism) and TSBMA (with the OP mechanism) on the instances $11Iy$ to $20Iy$.} 
	\begin{scriptsize}
	\setlength{\tabcolsep}{3mm}{
	\begin{tabular}{l|lll|lll}
	\toprule[0.75pt]

\multirow{2}{*}{Instance} & \multicolumn{3}{c|}{TSBMA$^-$} & \multicolumn{3}{c}{TSBMA} \\ 
\cmidrule(lr){2-4} \cmidrule(lr){5-7}

 & \textit{$f_{best}$} & \textit{$f_{avg}$} & \textit{$std$} 
 & \textit{$f_{best}$} & \textit{$f_{avg}$} & \textit{$std$}  \\
\hline

    $11I1$  & 4960  & 4960 & 0.00  & 4960  & 4960 & 0.00 \\
    $11I2$  & 4940  & 4940 & 0.00  & 4940  & 4940 & 0.00 \\
    $11I3$  & 4950  & 4949.45 & 2.18  & 4950  & \textbf{4950} & 0.00 \\
    $11I4$  & 4930  & 4924 & 4.42  & 4930  & \textbf{4930} & 0.00 \\
    $11I5$  & 4920  & 4916.35 & 4.68  & 4920  & \textbf{4920} & 0.00 \\
    $12I1$  & 4685  & 4676.95 & 4.99  & \textbf{4690} & \textbf{4687.65} & 2.22 \\
    $12I2$  & 4670  & 4668.70 & 3.10  & \textbf{4680} & \textbf{4680} & 0.00 \\
    $12I3$  & 4690  & 4685.45 & 4.20  & 4690  & \textbf{4690} & 0.00 \\
    $12I4$  & 4680  & 4669.80 & 6.36  & 4680  & \textbf{4679.50} & 2.18 \\
    $12I5$  & 4670  & 4664.50 & 4.57  & 4670  & \textbf{4670} & 0.00 \\
    $13I1$  & 4525  & 4511.20 & 8.55  & \textbf{4539} & \textbf{4534.80} & 3.60 \\
    $13I2$  & 4521  & 4509.25 & 7.29  & \textbf{4530} & \textbf{4528} & 4.00 \\
    $13I3$  & 4520  & 4515.40 & 4.55  & \textbf{4540} & \textbf{4531} & 3.00 \\
    $13I4$  & 4520  & 4507.10 & 6.94  & \textbf{4530} & \textbf{4529.15} & 2.29 \\
    $13I5$  & 4530  & 4513.65 & 6.51  & \textbf{4537} & \textbf{4534.20} & 3.43 \\
    $14I1$  & 4429  & 4413.55 & 7.41  & \textbf{4440} & \textbf{4440} & 0.00 \\
    $14I2$  & 4420  & 4413.55 & 4.47  & \textbf{4440} & \textbf{4439.40} & 0.49 \\
    $14I3$  & 4420  & 4415.20 & 4.70  & \textbf{4439} & \textbf{4439} & 0.00 \\
    $14I4$  & 4420  & 4412.40 & 4.57  & \textbf{4435} & \textbf{4431.50} & 2.06 \\
    $14I5$  & 4420  & 4413.85 & 4.27  & \textbf{4440} & \textbf{4440} & 0.00 \\
    $15I1$  & 4359  & 4346.15 & 5.06  & \textbf{4370} & \textbf{4369.95} & 0.22 \\
    $15I2$  & 4359  & 4344.10 & 6.22  & \textbf{4370} & \textbf{4370} & 0.00 \\
    $15I3$  & 4359  & 4341.85 & 6.54  & \textbf{4370} & \textbf{4369.25} & 1.84 \\
    $15I4$  & 4350  & 4341.05 & 7.78  & \textbf{4370} & \textbf{4369.85} & 0.36 \\
    $15I5$  & 4360  & 4346.10 & 5.47  & \textbf{4379} & \textbf{4373.15} & 4.29 \\
    $16I1$  & 5020  & 5013.75 & 4.93  & 5020  & \textbf{5020} & 0.00 \\
    $16I2$  & 5010  & 5003.30 & 5.60  & 5010  & \textbf{5010} & 0.00 \\
    $16I3$  & 5020  & 5010.65 & 5.33  & 5020  & \textbf{5020} & 0.00 \\
    $16I4$  & 5020  & 5008.95 & 8.24  & 5020  & \textbf{5020} & 0.00 \\
    $16I5$  & 5060  & 5052.85 & 8.37  & 5060  & \textbf{5060} & 0.00 \\
    $17I1$  & 4730  & 4707.50 & 7.51  & 4730  & \textbf{4729.70} & 0.64 \\
    $17I2$  & 4716  & 4704.50 & 6.27  & \textbf{4720} & \textbf{4719.50} & 2.18 \\
    $17I3$  & 4720  & 4705.10 & 6.68  & \textbf{4729} & \textbf{4723.60} & 4.41 \\
    $17I4$  & 4722  & 4701.20 & 9.68  & \textbf{4730} & \textbf{4730} & 0.00 \\
    $17I5$  & 4720  & 4706.20 & 8.37  & \textbf{4730} & \textbf{4726.85} & 4.50 \\
    $18I1$  & 4555  & 4539.75 & 6.31  & \textbf{4568} & \textbf{4565.80} & 3.40 \\
    $18I2$  & 4540  & 4532.20 & 4.64  & \textbf{4560} & \textbf{4551.40} & 3.01 \\
    $18I3$  & 4570  & 4545.20 & 8.58  & 4570  & \textbf{4569.40} & 2.20 \\
    $18I4$  & 4550  & 4539.30 & 6.75  & \textbf{4568} & \textbf{4565.20} & 3.12 \\
    $18I5$  & 4550  & 4542.50 & 5.32  & \textbf{4570} & \textbf{4567.95} & 3.46 \\
    $19I1$  & 4432  & 4424.65 & 4.71  & \textbf{4460} & \textbf{4456.65} & 3.48 \\
    $19I2$  & 4443  & 4430.85 & 6.06  & \textbf{4460} & \textbf{4453.25} & 4.17 \\
    $19I3$  & 4440  & 4428.15 & 6.01  & \textbf{4469} & \textbf{4462.05} & 4.04 \\
    $19I4$  & 4450  & 4431.25 & 5.63  & \textbf{4460} & \textbf{4453.20} & 3.89 \\
    $19I5$  & 4449  & 4435.65 & 5.42  & \textbf{4466} & \textbf{4460.75} & 1.61 \\
    $20I1$  & 4364  & 4358.95 & 2.80  & \textbf{4390} & \textbf{4383.20} & 3.36 \\
    $20I2$  & 4360  & 4356.85 & 4.25  & \textbf{4390} & \textbf{4381.80} & 3.78 \\
    $20I3$  & 4370  & 4360.45 & 5.11  & \textbf{4389} & \textbf{4387.90} & 2.77 \\
    $20I4$  & 4370  & 4359.75 & 5.78  & \textbf{4389} & \textbf{4380.40} & 1.98 \\
    $20I5$  & 4366  & 4357.45 & 4.78  & \textbf{4390} & \textbf{4386.40} & 4.05 \\
    \hline
    \#Avg & 4603.08 & 4593.13 & 5.56  & \textbf{4614.14} & \textbf{4611.83} & 1.80 \\
    \#Best & 15/50 & 2/50 & - & 50/50 & 50/50 & - \\
    $p$-$values$ & 2.51e-7 & 1.68e-9 & - & - & - & - \\

\bottomrule[0.75pt]
	\end{tabular}}

\label{Result_OP}
\end{scriptsize}
\end{table} 
\renewcommand{\baselinestretch}{1}\large\normalsize

TSBMA avoids revisiting previously encountered solutions with the OP mechanism introduced in Section \ref{Subsec_OP}. To assess the usefulness of the OP mechanism, we created a TSBMA variant (denoted by TSBMA$^-$) by disabling the OP component and keeping the other components unchanged. We ran TSBMA$^-$ to solve the 50 $11Iy$ to $20Iy$ instances according to experimental settings given in Section \ref{Subsec_Setting} and reported the results in Table \ref{Result_OP}. The first column gives the name of each instance and the remaining columns show the best objective values ($f_{best}$), the average objective values ($f_{avg}$) and the standard deviations ($std$). Row \#Avg presents the average value of each column and row \#Best indicates the number of instances for which an algorithm obtains the best values between the two sets of results. The last row shows the $p$-$values$ from the Wilcoxon signed-rank test. The best results of the compared algorithms are highlighted in bold. 

From Table \ref{Result_OP}, we observe that TSBMA$^-$ performs worse than TSMBA. TSBMA$^-$ obtains worse $f_{best}$ values for 35 out of the 50 instances and worse $f_{avg}$ values for 48 instances. Considering the $std$ values, TSBMA$^-$ shows a much less stable performance than TSMBA. Moreover, the small $p$-$values$ ($ < 0.05$) from the Wilcoxon tests confirm the statistically significant difference between the results of TSMBA and TSBMA$^-$. This experiment demonstrates the effectiveness and robustness of the operation-prohibiting mechanism employed by the TSMBA algorithm.

\section{Conclusions}
\label{Sec_Conclu}

The disjunctively constrained knapsack problem is a well-known NP-hard model. Given its practical significance and intrinsic difficulty, a variety of exact and heuristic algorithms have been designed for solving the problem. We proposed the threshold search based memetic algorithm that combines for the first time threshold search with the memetic framework.

Extensive evaluations on a large number of benchmark instances in the literature (6340 instances in total) showed that the algorithm performs competitively with respect to the state-of-the-art algorithms. Our approach is able to discover 24 new lower bounds out of the 100 instances of Set I and 354 new lower bounds out of the 6240 instances of Set II. These new lower bounds are useful for future studies on the DCKP. The algorithm also attains the best-known or known optimal results on most of the remaining instances. We carried out additional experiments to investigate the two essential ingredients of the algorithm (the threshold search technique and the operation-prohibiting mechanism). The disjunctively constrained knapsack problem is a useful model to formulate a number of practical applications. The algorithm and its code (that we will make available) can contribute to solving these problems. 

There are at least two possible directions for future work. First, TSBMA performed badly on most random sparse instances of $SR$. It would be interesting to improve the algorithm to better handle such instances. Second, given the good performance of the adopted approach, it is worth investigating its underlying ideas to solve related problems discussed in the introduction. 

\section*{Declaration of competing interest}
The authors declare that they have no known competing interests that could have appeared to influence the work reported in this paper.

\section*{Acknowledgments}
We would like to thank Dr. Zhe Quan, Dr. Lei Wu, Dr. Pablo San Segundo and their co-authors for sharing the instances of the DCKP and the detailed results of their algorithms reported in \cite{quan2017cooperative}, \cite{quan2017design}, and \cite{coniglio2020new}.

\bibliographystyle{elsart-num-sort}
%\bibliography{dckp}

\begin{appendix}

\section{Computational results on the 100 DCKP instances of Set I}
\label{Result_setI}

Tables \ref{Result_setI1} and \ref{Result_setI2} report the detailed computational results of the TSBMA algorithm and the reference algorithms (PNS \cite{quan2017design}, CPANS \cite{quan2017cooperative}  and PTS \cite{salem2017probabilistic}) on the 100 DCKP instances of Set I. 

The first two columns of the tables give the name of each instance and the best-known objective values (BKV) ever reported in the literature. We employ the following four performance indicators to present our results: best objective value ($f_{best}$), average objective value over 20 runs ($f_{avg}$), standard deviations over 20 runs ($std$), and average run time $t_{avg}$ in seconds to reach the best objective value. However, some of the performance indicators of the reference algorithms are not available in the literature (i.e., $f_{avg}$, $t_{avg}$ and $std$). Note that for \cite{quan2017design} (PNS) and \cite{quan2017cooperative} (CPANS), the authors reported several groups of results obtained by using different numbers of processors (range from 10 to 400). To make a fair comparison, we take the best $f_{best}$ value of each instance in these groups of results as the final result. We use the average of the $t_{avg}$ values in these groups as the final average run time. The last row \#Avg indicates the average value of each column. The 24 new lower bounds discovered by our TSBMA algorithm are highlighted in bold.

\renewcommand{\baselinestretch}{0.8}\large\normalsize
\begin{table}[!htbp]
\centering
	\caption{Computational results of the TSBMA algorithm with the reference algorithms on the 50 DCKP instances of Set I ($1Iy$ to $10Iy$).} 
	\begin{scriptsize}
	\setlength{\tabcolsep}{2.2mm}{
	\begin{tabular}{l|l|l|ll|ll|llll}
	\toprule[0.75pt]

\multirow{2}{*}{Instance} & \multirow{2}{*}{BKV} & PNS \cite{quan2017design} & \multicolumn{2}{c|}{CPANS \cite{quan2017cooperative}} & \multicolumn{2}{c|}{PTS \cite{salem2017probabilistic}} & \multicolumn{4}{c}{TSBMA (this work)} \\ 
\cmidrule(lr){3-3} \cmidrule(lr){4-5} \cmidrule(lr){6-7} \cmidrule(lr){8-11}

 &  & \textit{$f_{best}$} 
 & \textit{$f_{best}$} & \textit{$t_{avg} (s)$} & \textit{$f_{best}$} & \textit{$f_{avg}$}
 & \textit{$f_{best}$} & \textit{$f_{avg}$} & \textit{$std$} & \textit{$t_{avg} (s)$}  \\
\hline

    $1I1$  & 2567  & 2567  & 2567  & 17.133 & 2567  & 2567  & 2567  & 2567  & 0.00  & 163.577 \\
    $1I2$   & 2594  & 2594  & 2594  & 12.623 & 2594  & 2594  & 2594  & 2594  & 0.00  & 19.322 \\
    $1I3$   & 2320  & 2320  & 2320  & 14.897 & 2320  & 2320  & 2320  & 2320  & 0.00  & 6.060 \\
    $1I4$   & 2310  & 2310  & 2310  & 13.063 & 2310  & 2310  & 2310  & 2310  & 0.00  & 10.969 \\
    $1I5$   & 2330  & 2330  & 2330  & 20.757 & 2330  & 2321  & 2330  & 2330  & 0.00  & 63.663 \\
    $2I1$   & 2118  & 2118  & 2118  & 21.710 & 2118  & 2115.2 & 2118  & 2117.70 & 0.46  & 330.797 \\
    $2I2$  & 2118  & 2112  & 2118  & 129.390 & 2110  & 2110  & 2118  & 2111.60 & 3.20  & 705.755 \\
    $2I3$   & 2132  & 2132  & 2132  & 23.820 & 2119  & 2112.4 & 2132  & 2132  & 0.00  & 210.108 \\
    $2I4$   & 2109  & 2109  & 2109  & 31.377 & 2109  & 2105.6 & 2109  & 2109  & 0.00  & 14.182 \\
    $2I5$   & 2114  & 2114  & 2114  & 20.040 & 2114  & 2110.4 & 2114  & 2114  & 0.00  & 99.133 \\
    $3I1$   & 1845  & 1845  & 1845  & 34.683 & 1845  & 1760.3 & 1845  & 1845  & 0.00  & 3.780 \\
    $3I2$   & 1795  & 1795  & 1795  & 107.993 & 1795  & 1767.5 & 1795  & 1795  & 0.00  & 3.029 \\
    $3I3$   & 1774  & 1774  & 1774  & 22.490 & 1774  & 1757  & 1774  & 1774  & 0.00  & 3.585 \\
    $3I4$   & 1792  & 1792  & 1792  & 27.953 & 1792  & 1767.4 & 1792  & 1792  & 0.00  & 3.275 \\
    $3I5$   & 1794  & 1794  & 1794  & 34.820 & 1794  & 1755.5 & 1794  & 1794  & 0.00  & 9.159 \\
    $4I1$   & 1330  & 1330  & 1330  & 37.307 & 1330  & 1329.1 & 1330  & 1330  & 0.00  & 1.967 \\
    $4I2$   & 1378  & 1378  & 1378  & 40.827 & 1378  & 1370.5 & 1378  & 1378  & 0.00  & 3.926 \\
    $4I3$   & 1374  & 1374  & 1374  & 100.183 & 1374  & 1370  & 1374  & 1374  & 0.00  & 2.431 \\
    $4I4$   & 1353  & 1353  & 1353  & 26.930 & 1353  & 1337.6 & 1353  & 1353  & 0.00  & 4.167 \\
    $4I5$   & 1354  & 1354  & 1354  & 81.113 & 1354  & 1333.2 & 1354  & 1354  & 0.00  & 6.196 \\
    $5I1$   & 2700  & 2694  & 2700  & 122.637 & 2700  & 2697.9 & 2700  & 2700  & 0.00  & 78.215 \\
    $5I2$  & 2700  & 2700  & 2700  & 111.160 & 2700  & 2699  & 2700  & 2700  & 0.00  & 57.300 \\
    $5I3$   & 2690  & 2690  & 2690  & 73.640 & 2690  & 2689  & 2690  & 2690  & 0.00  & 18.566 \\
    $5I4$   & 2700  & 2700  & 2700  & 130.913 & 2700  & 2699  & 2700  & 2700  & 0.00  & 52.807 \\
    $5I5$   & 2689  & 2689  & 2689  & 279.377 & 2689  & 2682.7 & 2689  & 2687.65 & 3.21  & 289.966 \\
    $6I1$   & 2850  & 2850  & 2850  & 104.623 & 2850  & 2843  & 2850  & 2850  & 0.00  & 57.997 \\
    $6I2$   & 2830  & 2830  & 2830  & 93.887 & 2830  & 2829  & 2830  & 2830  & 0.00  & 76.883 \\
    $6I3$   & 2830  & 2830  & 2830  & 203.677 & 2830  & 2830  & 2830  & 2830  & 0.00  & 157.597 \\
    $6I4$   & 2830  & 2824  & 2830  & 160.587 & 2830  & 2824.7 & 2830  & 2830  & 0.00  & 328.817 \\
    $6I5$   & 2840  & 2831  & 2840  & 112.947 & 2840  & 2825  & 2840  & 2833.10 & 4.22  & 378.393 \\
    $7I1$   & 2780  & 2780  & 2780  & 186.970 & 2780  & 2771  & 2780  & 2779.40 & 1.43  & 483.465 \\
    $7I2$   & 2780  & 2780  & 2780  & 161.117 & 2780  & 2769.8 & 2780  & 2775.50 & 4.97  & 372.935 \\
    $7I3$   & 2770  & 2770  & 2770  & 136.310 & 2770  & 2762  & 2770  & 2768.50 & 3.57  & 393.018 \\
    $7I4$   & 2800  & 2800  & 2800  & 123.957 & 2800  & 2791.9 & 2800  & 2795.50 & 4.97  & 162.060 \\
    $7I5$   & 2770  & 2770  & 2770  & 149.933 & 2770  & 2763.6 & 2770  & 2770  & 0.00  & 290.591 \\
    $8I1$   & 2730  & 2720  & 2730  & 472.153 & 2720  & 2718.9 & 2730  & 2724  & 4.90  & 484.264 \\
    $8I2$   & 2720  & 2720  & 2720  & 109.373 & 2720  & 2713.6 & 2720  & 2720  & 0.00  & 214.760 \\
    $8I3$   & 2740  & 2740  & 2740  & 112.847 & 2740  & 2731.5 & 2740  & 2739.55 & 1.96  & 207.311 \\
    $8I4$   & 2720  & 2720  & 2720  & 253.230 & 2720  & 2712  & 2720  & 2715.35 & 4.85  & 518.579 \\
    $8I5$   & 2710  & 2710  & 2710  & 115.777 & 2710  & 2705  & 2710  & 2710  & 0.00  & 67.003 \\
    $9I1$  & 2680  & 2678  & 2680  & 134.023 & 2670  & 2666.9 & 2680  & 2679.70 & 0.71  & 316.210 \\
    $9I2$   & 2670  & 2670  & 2670  & 158.397 & 2670  & 2661.7 & 2670  & 2669.90 & 0.44  & 238.149 \\
    $9I3$   & 2670  & 2670  & 2670  & 123.280 & 2670  & 2666.5 & 2670  & 2670  & 0.00  & 161.176 \\
    $9I4$   & 2670  & 2670  & 2670  & 137.690 & 2663  & 2657.3 & 2670  & 2668.90 & 2.49  & 522.294 \\
    $9I5$   & 2670  & 2670  & 2670  & 131.247 & 2670  & 2662  & 2670  & 2670  & 0.00  & 98.124 \\
    $10I1$  & 2624  & 2620  & 2624  & 244.020 & 2620  & 2613.7 & 2624  & 2621.45 & 1.72  & 348.617 \\
    $10I2^*$  & 2642$^*$  & 2630  & 2630  & 144.867 & 2630  & 2620.8 & 2630  & 2630  & 0.00  & 182.474 \\
    $10I3$  & 2627  & 2620  & 2627  & 198.050 & 2620  & 2614.5 & 2627  & 2621.40 & 2.80  & 326.099 \\
    $10I4^*$  & 2621$^*$  & 2620  & 2620  & 148.997 & 2620  & 2609.7 & 2620  & 2620  & 0.00  & 105.609 \\
    $10I5$  & 2630  & 2627  & 2630  & 170.620 & 2627  & 2617.6 & 2630  & 2629.50 & 2.18  & 307.851 \\
    \hline
    \#Avg & 2403.68 & 2402.36 & 2403.42 & 112.508 & 2402.18 & 2393.26 & 2403.42 & 2402.47 & 0.96 & 179.244 \\
    
\bottomrule[0.75pt]
	\end{tabular}}

\label{Result_setI1}
\end{scriptsize}
\end{table} 
\renewcommand{\baselinestretch}{1}\large\normalsize

\renewcommand{\baselinestretch}{0.8}\large\normalsize
\begin{table}[!htbp]\centering
	\caption{Computational results and comparison of the TSBMA algorithm with the reference algorithms on the 50 DCKP instances of Set I ($11Iy$ to $20Iy$).} 
	\begin{scriptsize}
	\setlength{\tabcolsep}{2.2mm}{
	\begin{tabular}{l|l|l|ll|llll}
	\toprule[0.75pt]

\multirow{2}{*}{Instance} & \multirow{2}{*}{BKV} & PNS \cite{quan2017design} & \multicolumn{2}{c|}{CPANS \cite{quan2017cooperative}} & \multicolumn{4}{c}{TSBMA (this work)} \\ 
\cmidrule(lr){3-3} \cmidrule(lr){4-5} \cmidrule(lr){6-9}

 &  & \textit{$f_{best}$} & \textit{$f_{best}$} & \textit{$t_{avg} (s)$}
 & \textit{$f_{best}$} & \textit{$f_{avg}$} & \textit{$std$} & \textit{$t_{avg} (s)$}  \\
\hline

    $11I1$  & 4950  & 4950  & 4950  & 333.435 & \textbf{4960}  & 4960  & 0.00  & 4.594 \\
    $11I2$  & 4940  & 4940  & 4928  & 579.460 & 4940  & 4940  & 0.00  & 14.305 \\
    $11I3$  & 4925  & 4920  & 4925  & 178.400 & \textbf{4950}  & 4950  & 0.00  & 69.236 \\
    $11I4$  & 4910  & 4890  & 4910  & 320.067 & \textbf{4930}  & 4930  & 0.00  & 139.197 \\
    $11I5$  & 4900  & 4890  & 4900  & 222.053 & \textbf{4920}  & 4920  & 0.00  & 100.178 \\
    $12I1$  & 4690  & 4690  & 4690  & 230.563 & 4690  & 4687.65 & 2.22  & 416.088 \\
    $12I2$  & 4680  & 4680  & 4680  & 502.600 & 4680  & 4680  & 0.00  & 224.000 \\
    $12I3$  & 4690  & 4690  & 4690  & 229.116 & 4690  & 4690  & 0.00  & 215.103 \\
    $12I4$  & 4680  & 4680  & 4676  & 367.330 & 4680  & 4679.50 & 2.18  & 256.300 \\
    $12I5$  & 4670  & 4670  & 4670  & 487.563 & 4670  & 4670  & 0.00  & 79.190 \\
    $13I1$  & 4533  & 4533  & 4533  & 395.985 & \textbf{4539}  & 4534.80 & 3.60  & 415.880 \\
    $13I2$  & 4530  & 4530  & 4530  & 573.718 & 4530  & 4528  & 4.00  & 361.229 \\
    $13I3$  & 4540  & 4530  & 4540  & 901.620 & 4540  & 4531  & 3.00  & 498.622 \\
    $13I4$  & 4530  & 4530  & 4530  & 315.076 & 4530  & 4529.15 & 2.29  & 366.951 \\
    $13I5$  & 4537  & 4537  & 4537  & 343.240 & 4537  & 4534.20 & 3.43  & 425.064 \\
    $14I1$  & 4440  & 4440  & 4440  & 483.156 & 4440  & 4440  & 0.00  & 205.733 \\
    $14I2$  & 4440  & 4440  & 4440  & 735.505 & 4440  & 4439.40 & 0.49  & 438.190 \\
    $14I3$  & 4439  & 4439  & 4439  & 614.733 & 4439  & 4439  & 0.00  & 146.119 \\
    $14I4$  & 4435  & 4435  & 4434  & 533.908 & 4435  & 4431.50 & 2.06  & 106.389 \\
    $14I5$  & 4440  & 4440  & 4440  & 473.448 & 4440  & 4440  & 0.00  & 160.900 \\
    $15I1$  & 4370  & 4370  & 4370  & 797.125 & 4370  & 4369.95 & 0.22  & 321.296 \\
    $15I2$  & 4370  & 4370  & 4370  & 676.703 & 4370  & 4370  & 0.00  & 181.021 \\
    $15I3$  & 4370  & 4370  & 4370  & 612.792 & 4370  & 4369.25 & 1.84  & 315.575 \\
    $15I4$  & 4370  & 4370  & 4370  & 649.398 & 4370  & 4369.85 & 0.36  & 424.873 \\
    $15I5$  & 4379  & 4379  & 4379  & 678.354 & 4379  & 4373.15 & 4.29  & 359.003 \\
    $16I1$  & 4980  & 4980  & 4980  & 286.130 & \textbf{5020}  & 5020  & 0.00  & 205.964 \\
    $16I2$  & 4990  & 4990  & 4980  & 232.825 & \textbf{5010}  & 5010  & 0.00  & 342.824 \\
    $16I3$  & 5009  & 5000  & 5009  & 199.880 & \textbf{5020}  & 5020  & 0.00  & 155.070 \\
    $16I4$  & 5000  & 4997  & 5000  & 831.750 & \textbf{5020}  & 5020  & 0.00  & 86.324 \\
    $16I5$  & 5040  & 5020  & 5040  & 982.970 & \textbf{5060}  & 5060  & 0.00  & 32.837 \\
    $17I1$  & 4730  & 4730  & 4721  & 422.640 & 4730  & 4729.70 & 0.64  & 388.541 \\
    $17I2$  & 4710  & 4710  & 4710  & 248.770 & \textbf{4720}  & 4719.50 & 2.18  & 300.275 \\
    $17I3$  & 4720  & 4720  & 4720  & 454.317 & \textbf{4729}  & 4723.60 & 4.41  & 343.016 \\
    $17I4$  & 4720  & 4720  & 4720  & 432.900 & \textbf{4730}  & 4730  & 0.00  & 288.961 \\
    $17I5$  & 4720  & 4720  & 4720  & 102.468 & \textbf{4730}  & 4726.85 & 4.50  & 366.752 \\
    $18I1$  & 4566  & 4566  & 4566  & 225.010 & \textbf{4568}  & 4565.80 & 3.40  & 269.545 \\
    $18I2$  & 4550  & 4550  & 4550  & 288.862 & \textbf{4560}  & 4551.40 & 3.01  & 13.884 \\
    $18I3$  & 4570  & 4570  & 4570  & 328.555 & 4570  & 4569.40 & 2.20  & 466.748 \\
    $18I4$  & 4560  & 4560  & 4560  & 511.527 & \textbf{4568}  & 4565.20 & 3.12  & 264.931 \\
    $18I5$  & 4570  & 4570  & 4570  & 651.887 & 4570  & 4567.95 & 3.46  & 572.589 \\
    $19I1$  & 4460  & 4460  & 4460  & 506.945 & 4460  & 4456.65 & 3.48  & 459.570 \\
    $19I2$  & 4459  & 4459  & 4459  & 666.900 & \textbf{4460}  & 4453.25 & 4.17  & 307.224 \\
    $19I3$  & 4460  & 4460  & 4460  & 608.913 & \textbf{4469}  & 4462.05 & 4.04  & 485.550 \\
    $19I4$  & 4450  & 4450  & 4450  & 476.755 & \textbf{4460}  & 4453.20 & 3.89  & 430.824 \\
    $19I5$  & 4460  & 4460  & 4460  & 508.730 & \textbf{4466}  & 4460.75 & 1.61  & 40.752 \\
    $20I1$  & 4389  & 4389  & 4388  & 957.410 & \textbf{4390}  & 4383.20 & 3.36  & 929.372 \\
    $20I2$  & 4390  & 4390  & 4387  & 756.908 & 4390  & 4381.80 & 3.78  & 299.673 \\
    $20I3$  & 4389  & 4383  & 4389  & 966.010 & 4389  & 4387.90 & 2.77  & 568.988 \\
    $20I4$  & 4388  & 4388  & 4380  & 993.630 & \textbf{4389}  & 4380.40 & 1.98  & 657.694 \\
    $20I5$  & 4389  & 4389  & 4389  & 772.495 & \textbf{4390}  & 4386.40 & 4.05  & 646.570 \\
   	\hline
    \#Avg &	4608.54 & 4606.88 & 4607.58 & 513.011 & 4614.14 & 4611.83 & 1.80 & 303.390 \\

\bottomrule[0.75pt]
	\end{tabular}}

\label{Result_setI2}
\end{scriptsize}
\end{table} 
\renewcommand{\baselinestretch}{1}\large\normalsize

\end{appendix}
\end{document}